\documentclass[runningheads]{llncs}
\usepackage{graphicx}

\usepackage{tikz}
\usepackage{comment}
\usepackage{amsmath,amssymb} 
\usepackage{color}

\usepackage[breaklinks,colorlinks]{hyperref}

\usepackage[accsupp]{axessibility}  %

\begin{document}

\pagestyle{headings}
\mainmatter
\def\ECCVSubNumber{100}  

\title{ABAW: Learning from Synthetic Data \& Multi-Task Learning Challenges}


\titlerunning{ABAW: Learning from Synthetic Data \& Multi-Task Learning Challenges}

\author{Dimitrios Kollias
\authorrunning{D. Kollias}
\institute{Queen Mary University of London, UK \\
\email{d.kollias@qmul.ac.uk}}}

\maketitle

\begin{abstract}
This paper describes the fourth Affective Behavior Analysis in-the-wild (ABAW) Competition, held in conjunction with European Conference on Computer Vision (ECCV), 2022. The 4th ABAW Competition is a continuation of the Competitions held at IEEE CVPR 2022, ICCV 2021, IEEE FG 2020 and IEEE CVPR 2017 Conferences, and aims at automatically analyzing affect. In the previous runs of this Competition, the Challenges targeted Valence-Arousal Estimation, Expression Classification and  Action Unit Detection. This year the Competition encompasses two different Challenges: i) a Multi-Task-Learning one in which the goal is to learn at the same time (i.e., in a multi-task learning setting) all the three above mentioned tasks; and ii) a Learning from Synthetic Data one in which the goal is to learn to recognise the basic expressions from artificially generated data and generalise to real data.

The Aff-Wild2 database is a large scale in-the-wild database and the first one that contains annotations for valence and arousal, expressions and action units. This database is the basis for the above Challenges. In more detail: i) s-Aff-Wild2  -a static version of Aff-Wild2 database- has been constructed and utilized for the purposes of the Multi-Task-Learning Challenge; and ii) some specific frames-images from the Aff-Wild2 database have been used in an expression manipulation manner for creating the synthetic dataset, which is the basis for the Learning from Synthetic Data Challenge.
In this paper, at first we present the two Challenges, along with the utilized corpora, then we outline the evaluation metrics and finally present the baseline systems per Challenge, as well as their derived results. More information regarding the Competition can be found in the competition's website: \url{https://ibug.doc.ic.ac.uk/resources/eccv-2023-4th-abaw/}.

\end{abstract}

\section{Introduction}

Automatic facial behavior analysis has a long history of studies in the intersection of computer vision, physiology and psychology and has applications spread across a variety of fields, such as medicine, health, or driver fatigue, monitoring, e-learning, marketing, entertainment, lie detection and law. 
However it is only recently, with the collection of large-scale datasets and powerful machine learning methods such as deep neural networks, that automatic facial behavior analysis started to thrive.
When it comes to automatically recognising affect in-the-wild (i.e., in uncontrolled conditions and unconstrained environments), there exist three iconic tasks, which are: i) recognition of basic expressions
(anger, disgust, fear, happiness, sadness, surprise and the neutral state); ii) estimation of continuous affect (valence -how positive/negative a person is- and arousal -how active/passive a person is-); iii) detection of facial action units (coding of facial motion with respect to activation of facial muscles, e.g. upper/inner eyebrows, nose wrinkles).

Ekman \cite{ekman2002facial} defined the six basic emotions, i.e., Anger, Disgust, Fear, Happiness, Sadness, Surprise and the Neutral
State, based on a cross-culture study \cite{ekman2002facial}, which indicated that humans perceive certain basic emotions in the same way regardless of culture. Nevertheless, advanced research on neuroscience and psychology argued that the model of six basic emotions are culture-specific and not universal. Additionally, the affect model based on basic emotions is limited in the ability to represent the complexity and subtlety of our daily affective displays. Despite these findings, the categorical model that describes emotions in terms of discrete basic emotions is still the most popular perspective for Expression Recognition, due to its pioneering investigations along with the direct and intuitive definition of facial expressions.

The dimensional model of affect, that is appropriate to represent not only extreme, but also subtle emotions
appearing in everyday human-computer interactions, has also attracted significant attention over the last years.
According to the dimensional approach \cite{frijda1986emotions,whissel1989dictionary,russell1978evidence}, affective behavior is described by a number of latent continuous
dimensions. The most commonly used dimensions include valence (indicating how positive or negative an emotional
state is) and arousal (measuring the power of emotion activation).

Detection of Facial Action Units (AUs) has also attained large attention. The Facial Action Coding System
(FACS) \cite{ekman2002facial,darwin1998expression} provides a standardised taxonomy of facial muscles’ movements and has been widely adopted as a common standard towards systematically categorising physical manifestation of complex facial expressions. Since any facial expression can be represented as a combination of action units, they constitute a natural physiological basis for face analysis. Consequently, in the last
years, there has been a shift of related research towards the detection of action units. The presence of action units
is typically brief and unconscious, and their detection requires analyzing subtle appearance changes in the human
face. Furthermore, action units do not appear in isolation, but as elemental units of facial expressions, and hence
some AUs co-occur frequently, while others are mutually exclusive.

The fourth Affective Behavior Analysis
in-the-wild (ABAW) Competition, held in conjunction with the European
Conference on Computer Vision (ECCV), 2022,  is a continuation of the first \footnote{\url{https://ibug.doc.ic.ac.uk/resources/fg-2020-competition-affective-behavior-analysis/}} \cite{kollias2020analysing}, second \footnote{\url{https://ibug.doc.ic.ac.uk/resources/iccv-2021-2nd-abaw/}} \cite{kollias2021analysing} and third \cite{kollias2022abaw} \footnote{\url{https://ibug.doc.ic.ac.uk/resources/cvpr-2022-3rd-abaw/}}  ABAW Competitions held in conjunction with  the IEEE  Conference  on  Face  and  Gesture Recognition (IEEE FG) 2021, with the International Conference on Computer Vision (ICCV) 2022 and the IEEE International Conference on Computer Vision and Pattern Recognition (CVPR) 2022, respectively. The previous Competitions targeted dimensional (in terms of valence and arousal) \cite{deng2020multitask,zhang2020m,do2020affective,deng2021towards,li2021technical,zhang2021prior,vu2021multitask,wang2021multi,zhang2021audio,xie2021technical} \cite{jin2021multi,antoniadis2021audiovisual,oh2021causal,Zhang_2022_CVPR,Karas_2022_CVPR,Nguyen_2022_CVPR,meng2022multi,zhang2022continuous,nguyen2022ensemble,savchenko2022frame,karas2022continuous,rajasekar2022joint,zhang2022transformer,Meng_2022_CVPR}, categorical (in terms of the basic expressions) \cite{kuhnke2020two,gera2020affect,dresvyanskiy2020audio,youoku2020multi,liu2020emotion,gera2021affect,mao2021spatial,zhang2022transformer,jeong2022facial,Zhang_2021_ICCV,xue2022coarse,savchenko2022frame,phan2022expression,kim2022facial,yu2022multi,Jeong_2022_CVPR,Nguyen_2022_CVPR} and facial action unit analysis and recognition \cite{pahl2020multi,ji2020multi,han2016incremental,deng2020fau,saito2021action,vu2021multitask,zhang2022transformer,jiang2022facial,nguyen2022ensemble,wang2022multi,savchenko2022frame,tallec2022multi,hoai2022attention,wang2022facial,Jiang_2022_CVPR,Le_Hoai_2022_CVPR,Wang_2022_CVPR}. The third ABAW Challenge further targeted Multi-Task Learning for valence and arousal estimation, expression recognition and action unit detection \cite{deng2022multiple,jeong2022multitask,savchenko2022frame} \cite{zhang2022transformer,Deng_2022_CVPR,Jeong_2022_CVPR,Savchenko_2022_CVPR,zhang2022transformer}.

The fourth ABAW Competition contains two Challenges (i) the Multi-Task-Learning (MTL) one in which the goal is to create a system that learns at the same time (i.e., in a multi-task learning
setting) to estimate valence and arousal, classify eight expressions (6 basic expressions plus the neutral state plus a category 'other' which denotes expressions/affective states other than the 6 basic ones) and detect twelve action units; ii) the Learning from Synthetic Data (LSD) one in which the goal is to create a system that learns to recognise the six basic expressions (anger, disgust, fear, happiness, sadness, surprise) from artificially generated data (i.e., synthetic data) and generalise its knowledge to real-world (i.e., real) data.

Both Challenges' corpora are based on the Aff-Wild2 database \cite{kollias2022abaw,kollias2021analysing,kollias2020analysing,kolliasexpression,kollias2021affect} \cite{kollias2018aff2,kollias2018multi,kollias2021distribution,kollias2019face,kollias2018deep,zafeiriou,zafeiriou1}, which is the first comprehensive in-the-wild benchmark for all the three above-mentioned affect recognition tasks; the Aff-Wild2 database is an extensions of the Aff-Wild database \cite{kollias2018deep,zafeiriou,zafeiriou1}, with more videos and annotations for all behavior tasks. The MTL Challenge utilises a a static version of the Aff-Wild2 database, named s-Aff-Wild2. The LSD Challenge utilizes a synthetic dataset which has been constructed after manipulating the displayed expressions in some frames of the Aff-Wild2 database.

The remainder of this paper is organised as follows. The Competition corpora is introduced in Section \ref{corpora}, the Competition evaluation metrics are mentioned and described in Section \ref{metrics}, the developed baselines in each Challenge are explained and their obtained results are presented in Section \ref{baseline}, before concluding in Section \ref{conclusion}.


\section{Competition Corpora}\label{corpora}

The fourth Affective Behavior Analysis in-the-wild (ABAW) Competition relies on the Aff-Wild2 database, which is the first ever database annotated in terms of the tasks of: valence-arousal estimation, action unit detection and expression recognition. These three tasks constitute the basis of the two Challenges.


In the following, we provide a short overview of each Challenge's dataset along with a description of the pre-processing steps that we carried out for cropping and/or aligning the images of Aff-Wild2. These images have been utilized in our baseline experiments.

\subsection{Multi-Task Learning Challenge}

A static version of the Aff-Wild2 database has been generated by selecting some specific frames of the database; this Challenge's corpora is named s-Aff-Wild2. 
In total, 221,928 images are used that contain annotations in terms of: i) valence and arousal; ii) 6 basic expressions (anger, disgust, fear, happiness, sadness, surprise), plus the neutral state, plus the 'other' category (which denotes expressions/affective states other than the 6 basic ones); 12 action units.  

\begin{figure}[!h]
\centering
\includegraphics[height=6.5cm]{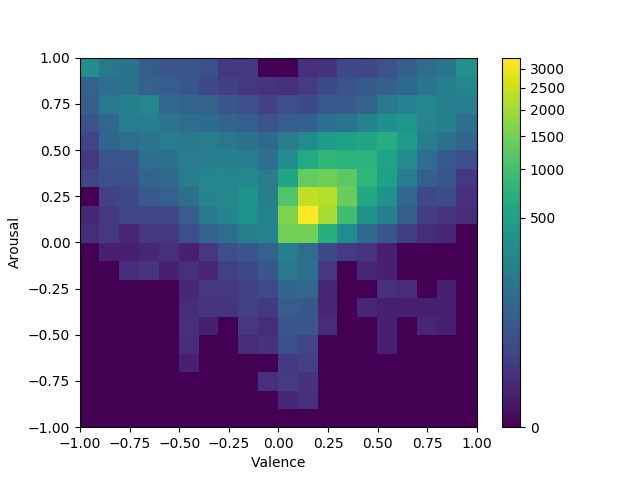}
\caption{Multi-Task-Learning Challenge: 2D Valence-Arousal Histogram of Annotations in s-Aff-Wild2}
\label{va_annot}
\end{figure}

Figure \ref{va_annot} shows the 2D Valence-Arousal histogram of annotations of s-Aff-Wild2.
Table \ref{expr_distr} shows the distribution of the 8 expression annotations of s-Aff-Wild2.
Table \ref{au_distr} shows the name of the 12 action units that have been annotated, the action that they correspond to and the distribution of their annotations in s-Aff-Wild2.

The s-Aff-Wild2 database is split into training, validation and test sets. At first the training and validation sets, along with their corresponding annotations, are being made public to the participants, so that they can develop their own methodologies and test them. At a later stage, the test set without annotations is given to the participants.

The participants are given two versions of s-Aff-Wild2: the cropped and cropped-aligned ones. At first, all images/frames of s-Aff-Wild2 are passed through the RetinaFace detector \cite{deng2020retinaface} so as to extract, for each image/frame, face bounding boxes and 5 facial landmarks. The images/frames are then cropped according the bounding box locations. All cropped-aligned images have the same dimensions $112 \times 112 \times 3$. These cropped  images/frames constitute the cropped version of s-Aff-Wild2 that is given to the participants. 
The 5 facial landmarks (two eyes, nose and two mouth corners) have then been used to perform similarity transformation. The resulting cropped-aligned images/frames constitute the cropped-aligned version of s-Aff-Wild2 that is given to the participants. The cropped-aligned version has been utilized in our baseline experiments, described in Section \ref{baseline}.

\begin{table}[!h]
\caption{Multi-Task-Learning Challenge:  Number of Annotated Images for each of the 8 Expressions}
\label{expr_distr}
\centering
\begin{tabular}{ |c||c| }
\hline
 Expressions & No of Images \\
\hline
\hline
Neutral & 37,073  \\
 \hline
Anger & 8,094  \\
 \hline
Disgust & 5,922 \\
 \hline
Fear &  6,899 \\
 \hline
Happiness & 32,397  \\
 \hline
Sadness & 13,447  \\
 \hline
Surprise & 9,873  \\
 \hline
 Other & 39,701 \\
 \hline
\end{tabular}
\end{table}

\begin{table}[!h]
    \centering
        \caption{Multi-Task-Learning Challenge: : Distribution of AU Annotations in Aff-Wild2}
    \label{au_distr}
\begin{tabular}{|c|c|c|}
\hline
  Action Unit \# & Action   &\begin{tabular}{@{}c@{}} Total Number \\  of Activated AUs \end{tabular} \\   \hline    
    \hline    
   AU 1 & inner brow raiser   & 29,995 \\   \hline    
   AU 2 & outer brow raiser  & 14,183 \\   \hline   
   AU 4 & brow lowerer   & 31,926 \\  \hline    
   AU 6 & cheek raiser  & 49,413 \\  \hline    
   AU 7 & lid tightener  & 72,806 \\  \hline    
   AU 10 & upper lip raiser  & 68,090 \\  \hline    
   AU 12 & lip corner puller  & 47,820 \\  \hline    
   AU 15 & lip corner depressor  & 5,105 \\  \hline   
  AU 23 & lip tightener & 6,538 \\  \hline    
   AU 24 & lip pressor & 8,052 \\  \hline    
   AU 25 & lips part  & 122,518 \\  \hline     
   AU 26 & jaw drop  & 19,439 \\  \hline     
\end{tabular}
\end{table}

Let us note that for the purposes of this Challenge, all participants are allowed to use the provided s-Aff-Wild2 database and/or any publicly available or private database; the participants are not allowed to use the audiovisual (A/V) Aff-Wild2 database (images and annotations).
Any methodological solution will be accepted for this Challenge.

\subsection{Learning from Synthetic Data Challenge}

Some specific cropped images/frames of the Aff-Wild2 database have been selected; these images/frames, which show a face with an arbitrary expression/affective state, have been used -in a facial expression manipulation manner \cite{kollias2018photorealistic,kolliasijcv,kollias2020va}- so as to synthesize basic facial expressions of the same person. Therefore a synthetic facial dataset has been generated and used for the purposes of this Challenge.
In total, 277,251 images that contain annotations in terms of the 6 basic expressions (anger, disgust, fear, happiness, sadness, surprise) have been generated. These images constitute the training set of this Challenge.
Table \ref{expr_distr2} shows the distribution of the 6 basic expression annotations of these generated images.
The validation and test sets of this Challenge are real images of the Aff-Wild2 database.
Let us note that the synthetic data have been generated from subjects of the validation set, but not of the test set.

At first the training (synthetic data) and validation (real data) sets, along with their corresponding annotations, are being made public to the participants, so that they can develop their own methodologies and test them. At a later stage, the test set (real data) without annotations is given to the participants.

Let us note that for the purposes of this Challenge, all participants are allowed to use any -publicly or not- available pre-trained model (as long as it has not been pre-trained on Aff-Wild2). The pre-trained model can be pre-trained on any task (eg VA estimation, Expression Classification, AU detection, Face Recognition). However when the teams are refining the model and developing the methodology they must only use the provided synthetic data. No real data should be used in model training/methodology development.

\begin{table}[!h]
\caption{Learning from Synthetic Data Challenge:  Number of Annotated Images for each of the 6 basic Expressions}
\label{expr_distr2}
\centering
\begin{tabular}{ |c||c| }
\hline
 Expressions & No of Images \\
\hline
\hline
Anger & 18,286  \\
 \hline
Disgust & 15,150 \\
 \hline
Fear & 10,923  \\
 \hline
Happiness & 73,285  \\
 \hline
Sadness & 144,631  \\
 \hline
Surprise & 14,976  \\
 \hline
\end{tabular}
\end{table}

\section{Evaluation Metrics for each Challenge}\label{metrics}

Next, we present the metrics that will be used for assessing the performance of the developed methodologies of the participating teams in each Challenge.

\subsection{Multi-Task Learning Challenge} 

The performance measure is the sum of: the average between the Concordance Correlation Coefficient (CCC) of valence and arousal; the average F1 Score of the 8 expression categories (i.e., macro F1 Score); the average F1 Score of the 12 action units (i.e., macro F1 Score).

CCC takes values in the range $[-1,1]$; high values are desired. CCC is defined as follows:

\begin{equation} \label{ccc}
\rho_c = \frac{2 s_{xy}}{s_x^2 + s_y^2 + (\bar{x} - \bar{y})^2},
\end{equation}

\noindent
where $s_x$ and $s_y$ are the variances of all video valence/arousal annotations and predicted values, respectively, $\bar{x}$ and $\bar{y}$ are their corresponding mean values and $s_{xy}$ is the corresponding covariance value.

The $F_1$ score is a weighted average of the recall (i.e., the ability of the classifier to find all the positive samples) and precision (i.e., the ability of the classifier not to label as positive a sample that is negative). The $F_1$ score  takes values in the range $[0,1]$; high values are desired. The $F_1$ score is defined as:

\begin{equation} \label{f1}
F_1 = \frac{2 \times precision \times recall}{precision + recall}
\end{equation}

Therefore, the evaluation criterion for the  Multi-Task-Learning Challenge is:

\begin{align} \label{mtll}
\mathcal{P}_{MTL} &= \mathcal{P}_{VA} + \mathcal{P}_{EXPR} + \mathcal{P}_{AU} \nonumber \\
&=  \frac{\rho_a + \rho_v}{2} + \frac{\sum_{expr} F_1^{expr}}{8} + \frac{\sum_{au} F_1^{au}}{12}
\end{align}

\subsection{Learning from Synthetic Data Challenge}

The performance measure is the average F1 Score of the 6 basic expression categories (i.e., macro F1 Score):

\begin{align} \label{lsd}
\mathcal{P}_{LSD} &=  \frac{\sum_{expr} F_1^{expr}}{6}
\end{align}

\section{Baseline Networks and Performance} \label{baseline}

All baseline systems rely exclusively on existing open-source machine learning toolkits to ensure the reproducibility of the results. All systems have been implemented in TensorFlow; training time was around five hours on a Titan X GPU, with a learning rate of $10^{-4}$ and with a batch size of 128.

In this Section, we first describe the baseline systems developed for each Challenge and then report their achieved performance.

\subsection{Multi-Task Learning Challenge}

The baseline network is a VGG16 network with with fixed convolutional weights (only the 3 fully connected layers were trained), pre-trained on the VGGFACE dataset. The output layer consists of 22 units: 2 linear units that give the valence and arousal predictions; 8 units equipped with softmax activation function that give the expression predictions; 12 units equipped with sigmoid activation function that give the action unit predictions. 

Let us mention here that no data augmentation techniques \cite{psaroudakis2022mixaugment} have been utilized when training this baseline network with the cropped-aligned version of s-Aff-Wild2 database. We just normalised all images' pixel intensity values in the range $[-1,1]$.

Table \ref{mtl:results} illustrates the performance of the baseline model on the validation set of s-Aff-Wild2.

\begin{table}[!h]
\caption{Multi-Task Learning Challenge:  Performance of baseline model on the validation set; evaluation criterion is the sum of each task’s independent
performance metric.}
\label{mtl:results}
\centering
\begin{tabular}{ |c||c| }
\hline
 Baseline & $\mathcal{P}_{MTL}$ \\
\hline
\hline
VGGFACE & 0.30  \\
 \hline
\end{tabular}
\end{table}

\subsection{Learning from Synthetic Data Challenge}

The baseline network is a ResNet with 50 layers, pre-trained on ImageNet (ResNet50); its output layer consists of 6 units and is equipped with softmax activation function that gives the basic expression predictions. 

Let us mention here that no data augmentation techniques have been utilized when training this baseline network with the synthetic images. We just normalised all images' pixel intensity values in the range $[-1,1]$.

Table \ref{lsd:results} illustrates the performance of the baseline model on the validation and test sets, which consist of only real data of the Aff-Wild2 database.

\begin{table}[!h]
\caption{Learning from Synthetic Data Challenge:  Performance of baseline model on the validation and test sets, which consist of only real data of the Aff-Wild2 database; evaluation criterion is the average F1 Score of the 6 basic expression
categories. The performance on the validation set is indicated inside the parenthesis.}
\label{lsd:results}
\centering
\begin{tabular}{ |c||c| }
\hline
 Baseline & $\mathcal{P}_{LSD}$ \\
\hline
\hline
ResNet50 & 0.30 (0.50)  \\
 \hline
\end{tabular}
\end{table}

\section{Conclusion}\label{conclusion}
In this paper we have presented the fourth Affective Behavior Analysis in-the-wild Competition (ABAW) 2022 held in conjunction with ECCV 2022. This Competition is a continuation of the first, second and third ABAW Competitions held in conjunction with  IEEE FG 2020, ICCV 2021 and IEEE CVPR 2022, respectively.  This Competition comprises two Challenges: i) the Multi-Task-
Learning (MTL) Challenge in which the goal is to create a system that learns at
the same time (i.e., in a multi-task learning setting) to estimate valence and
arousal, classify eight expressions (6 basic expressions plus the neutral state
plus a category ’other’ which denotes expressions/affective states other than the
6 basic ones) and detect twelve action units; ii) the Learning from Synthetic
Data (LSD) Challenge in which the goal is to create a system that learns to recognise
the six basic expressions (anger, disgust, fear, happiness, sadness, surprise) from
artificially generated data (i.e., synthetic data) and generalise its knowledge to
real-world (i.e., real) data.
Each Challenge's corpora is derived from the Aff-Wild2 database.

\clearpage

\bibliographystyle{splncs04}
\bibliography{egbib}
\end{document}